\documentclass{article}

\usepackage{microtype}
\usepackage{graphicx}
\usepackage{subfigure}
\usepackage{xcolor}
\usepackage{booktabs} 
\usepackage{multirow}

\usepackage{pgfplots}
\usetikzlibrary{pgfplots.groupplots}
\pgfplotsset{compat=1.3}
\usepackage{tikz}
\usetikzlibrary{patterns}
\usepackage[most]{tcolorbox}

\definecolor{battleshipgrey}{rgb}{0.3, 0.3, 0.3}
\definecolor{brilliantrose}{rgb}{1.0, 0.33, 0.64}
\definecolor{americanrose}{rgb}{1.0, 0.01, 0.24}
\definecolor{jweigreen}{rgb}{0,0.45,0.24}
\definecolor{bluegray}{rgb}{0.1, 0.1, 0.4}
\definecolor{ao(english)}{rgb}{0.0, 0.5, 0.0}
\definecolor{blanchedalmond}{rgb}{1.0, 0.92, 0.8}
\definecolor{atomictangerine}{rgb}{1.0, 0.6, 0.4}
\definecolor{chocolate(web)}{rgb}{0.82, 0.41, 0.12}
\definecolor{bananayellow}{rgb}{1.0, 0.88, 0.21}
\definecolor{goldenbrown}{rgb}{0.6, 0.4, 0.08}
\definecolor{aliceblue}{rgb}{0.94, 0.97, 1.0}
\definecolor{beige}{rgb}{0.96, 0.96, 0.86}
\definecolor{babyblue}{rgb}{0.54, 0.81, 0.94}
\definecolor{camel}{rgb}{0.76, 0.6, 0.42}
\definecolor{cinnamon}{rgb}{0.82, 0.41, 0.12}
\definecolor{deepskyblue}{rgb}{0.0, 0.75, 1.0}
\definecolor{frenchblue}{rgb}{0.0, 0.45, 0.73}
\definecolor{classicrose}{rgb}{0.98, 0.8, 0.91}
\definecolor{frenchrose}{rgb}{0.96, 0.29, 0.54}
\definecolor{frenchlilac}{rgb}{0.53, 0.38, 0.56}
\definecolor{frenchbeige}{rgb}{0.65, 0.48, 0.36}

\definecolor{forestgreen}{HTML}{2e7d43}
\definecolor{color1}{HTML}{FF9999}
\definecolor{color2}{HTML}{FF6666}
\definecolor{color3}{HTML}{FF3333}
\definecolor{color4}{HTML}{E60000}
\definecolor{color5}{HTML}{B30000}
\definecolor{color6}{HTML}{8CD98C}
\definecolor{color7}{HTML}{53c653}
\definecolor{color8}{HTML}{39ac39}
\definecolor{color9}{HTML}{2d862d}
\definecolor{color10}{HTML}{206020}
\definecolor{color11}{HTML}{cca300}

\usepackage{hyperref}

 \usepackage[accepted]{icml2024}

\icmltitlerunning{Agent-FLAN: Designing Data and Methods of Effective Agent Tuning for Large Language Models}

\begin{document}

\twocolumn[
\icmltitle{Agent-FLAN: Designing Data and Methods of Effective Agent Tuning \\for Large Language Models}




\begin{icmlauthorlist}
\icmlauthor{Zehui Chen}{ustc}
\icmlauthor{Kuikun Liu}{ailab}
\icmlauthor{Qiuchen Wang}{ustc}
\icmlauthor{Wenwei Zhang}{ailab}
\icmlauthor{Jiangning Liu}{ailab}\\
\icmlauthor{Dahua Lin}{ailab}
\icmlauthor{Kai Chen}{ailab}
\icmlauthor{Feng Zhao}{ustc}
\end{icmlauthorlist}

\icmlaffiliation{ustc}{Department of Automation, University of Science and Technology of China}
\icmlaffiliation{ailab}{Shanghai AI Labortary}

\icmlcorrespondingauthor{Kai Chen, Feng Zhao}{chenkai@pjlab.org.cn, fzhao956@ustc.edu.cn}

\icmlkeywords{Machine Learning, ICML}
\vskip 0.3in
]



\printAffiliationsAndNotice{}  
\setcounter{footnote}{1}%

\begin{abstract}
Open-sourced Large Language Models (LLMs) have achieved great success in various NLP tasks, however, they are still far inferior to API-based models when acting as agents. How to integrate agent ability into general LLMs becomes a crucial and urgent problem.
This paper first delivers three key observations: (1) the current agent training corpus is entangled with both formats following and agent reasoning, which significantly shifts from the distribution of its pre-training data; (2) LLMs exhibit different learning speeds on the capabilities required by agent tasks; and (3) current approaches have side-effects when improving agent abilities by introducing hallucinations. Based on the above findings, we propose Agent-FLAN to effectively Fine-tune LANguage models for Agents.
Through careful decomposition and redesign of the training corpus, Agent-FLAN enables Llama2-7B to outperform prior best works by 3.5\% across various agent evaluation datasets. With comprehensively constructed negative samples, Agent-FLAN greatly alleviates the hallucination issues based on our established evaluation benchmark. Besides, it consistently improves the agent capability of LLMs when scaling model sizes while slightly enhancing the general capability of LLMs. The code will be available at \href{https://github.com/InternLM/Agent-FLAN}{https://github.com/InternLM/Agent-FLAN}.
\end{abstract}

\input{figures/ablations-figure-2}

\section{Introduction}

Language agent \cite{mialon2023augmented,sumers2023cognitive}, which leverages the powerful abilities of LLMs to perceive environments, make decisions, and take actions, has emerged as an effective solution to complicated real-world problems \cite{song2023llm,huang2023inner}. Existing studies mainly focus on either prompt engineering \cite{wei2022chain,yao2022react,shinn2023reflexion} or framework scheduling of one/multiple close-sourced LLM (GPT-4) \cite{wu2023autogen,talebirad2023multi,hong2023metagpt} to accomplish agent tasks. Despite the impressive results and flexibility of these works, close-sourced LLMs suffer from steep financial cost and safety issues \cite{li2023multi,yuan2023gpt}, impeding them from further deployment.

Recently, open-sourced LLMs emerged as effective alternatives and show promising results in various applications \cite{vicuna2023,touvron2023llama,jiang2024mixtral}. However, a lingering challenge persists: while these LLMs excel in tasks necessitating linguistic finesse, they fall short when deployed as agents, especially when compared to API-based models \cite{zeng2023agenttuning,liu2023agentbench}. Efforts have been made to fine-tune LLMs for specific vertical fields \cite{qin2023toolllm,gou2023tora}, but this approach risks compromising their general capabilities. Thus, integrating effective agent abilities into general LLMs becomes a critical and pressing challenge in the field, aiming to bridge the gap between specialized and versatile language models.

In this paper, we embark on a journey to mitigate this disparity between the prowess of open-sourced LLMs and the superior performance exhibited by API-based models in agent-centric tasks. We first unravel three critical observations that serve as the foundation for our exploration: (1) most agent training data is entangled with both format following and general reasoning, deviating from the original pretrain domain of the model, \textit{i.e.,} natural conversation, leading to inadequate agent learning, (2) by explicitly decomposing the training data along the basic capabilities, each loss exhibits different convergence curves, indicating varied learning speeds on the abilities requisite for agent tasks, and (3) the existing approaches that predominantly concentrate on specific agent abilities, overlooking the prevalence and significance of hallucinations in the model's output.

Building upon these key insights, we present Agent-FLAN, which investigates the designing choice of effective \textbf{Agent} {\textbf{F}inetuning for large \textbf{LAN}guage models. Specifically, we tease apart format following and common reasoning in the agent training corpus, aligning the fine-tuning process to the pretrained domain of the language model. This elicits the pure agent abilities in LLMs without overfitting to specific format protocols. By further disentangling the agent tasks into distinct facets along the fundamental competencies of LLMs, Agent-FLAN affords training flexibilities depending on the varying learning rates of respective competencies. To comprehensively resolve the hallucination issues in agent tasks, we construct the \textit{Agent-H} benchmark, which assesses the hallucination issues of LLMs from various aspects. Subsequently, we meticulously curate diverse `negative' training samples to mitigate this problem effectively. 

We employ Agent-FLAN on the open-sourced Llama2-series, which surpasses prior works by a substantial 3.5\% margin across a spectrum of agent evaluation benchmarks, including general agent tasks and tool utilization. Furthermore, we provide a deeper understanding of the dynamics involved in agent tuning: the scaling laws governing data and model dimensions, and the intricate relationships between general and agent-specific tasks.
Our major contributions are as follows:
\begin{itemize}
	\item We identify three critical observations that hinder open-sourced LLMs from achieving competitive performance in the agent domain, offering valuable insights into the complex landscape of agent tuning.
	\item Based on above findings, we introduce Agent-FLAN, an innovative approach aiming to integrate effective agent abilities into general LLMs: aligning agent tuning to chat format (\S \ref{sec:method_align}), capabilities decomposition and data balancing (\S \ref{sec:method_decompose}), and negative sample constructions for hallucination elimination (\S \ref{sec:method_negative}). 
	\item Agent-FLAN outperforms prior works by a substantial 3.5\% margin on Llama2-series across a spectrum of agent evaluation benchmarks. Besides, we further study the dynamics of agent tuning, including scaling laws in terms of data and model scales, and intricate relationships between general and agent-specific tasks.
\end{itemize}

\section{Related Work}

 \subsection{LLM as Agent}

The emergence of large language models (LLMs) represents a significant stride towards Artificial General Intelligence (AGI). With the advancement of LLMs, language agents built upon LLMs engage with the world to accomplish a diverse array of tasks, having become a focal point of research \cite{wang2023survey,xi2023rise}. LLMs have been utilized in various studies for specific agent tasks, including web browsing\cite{deng2023mind2web,zhou2023webarena}, online shopping \cite{yao2022webshop}, database operations\cite{liu2023agentbench}, science experiments \cite{wang2022scienceworld}, observation free reasoning \cite{xu2023rewoo}, Wikipedia Q\&A \cite{yang2018hotpotqa}, daily computer tasks \cite{kim2023language} and household exploration \cite{shridhar2020alfworld}. 

In addition to research focused on specific tasks, there are ongoing studies concerning AI agents based on LLMs. Some studies, such as ReAct\cite{yao2022react}, emphasize actions during thinking, leading to significant improvements over various methods. While other works focus primarily on human and social property inside agents \cite{mao2023editing,park2023generative,zhang2023exploring}, intelligence collaboration within multiple agents \cite{chen2023agentverse,liu2023bolaa,liang2023encouraging}. Different from above, Agent-FLAN facilitates the integration of effective agent capabilities into general LLMs, enabling models to better understand and tackle complex problems in the real world.

 \subsection{Language Model Fine-Tuning}

Language Model Fine-Tuning is also a research hotspot, involving the adjustment of pre-trained models to adapt to specific tasks, aiming to align the output with expectations \cite{zhang2023instruction}. Various research studies have been conducted on fine-tuning to optimize the model’s reasoning capabilities \cite{liu2021generated,fu2023specializing}, proficiency in tools \cite{patil2023gorilla,qin2023toolllm,qiao2023making}, planning capability \cite{chen2023fireact}, retrieval-augmented \cite{wang2023instructuie}, etc. Additionally, there are various studies conducted on fine-tuning methods \cite{hu2021lora,ivison2022hint,dettmers2023qlora,lv2023full}, data selection principles \cite{gunasekar2023textbooks} and fine-tuning datasets \cite{sanh2021multitask,wang2022self,honovich2022unnatural,longpre2023flan}. 
\section{Pilot Observations}
\label{sec:pilot}
\begin{figure}[!t]
    \centering    
    \includegraphics[width=0.95\columnwidth]{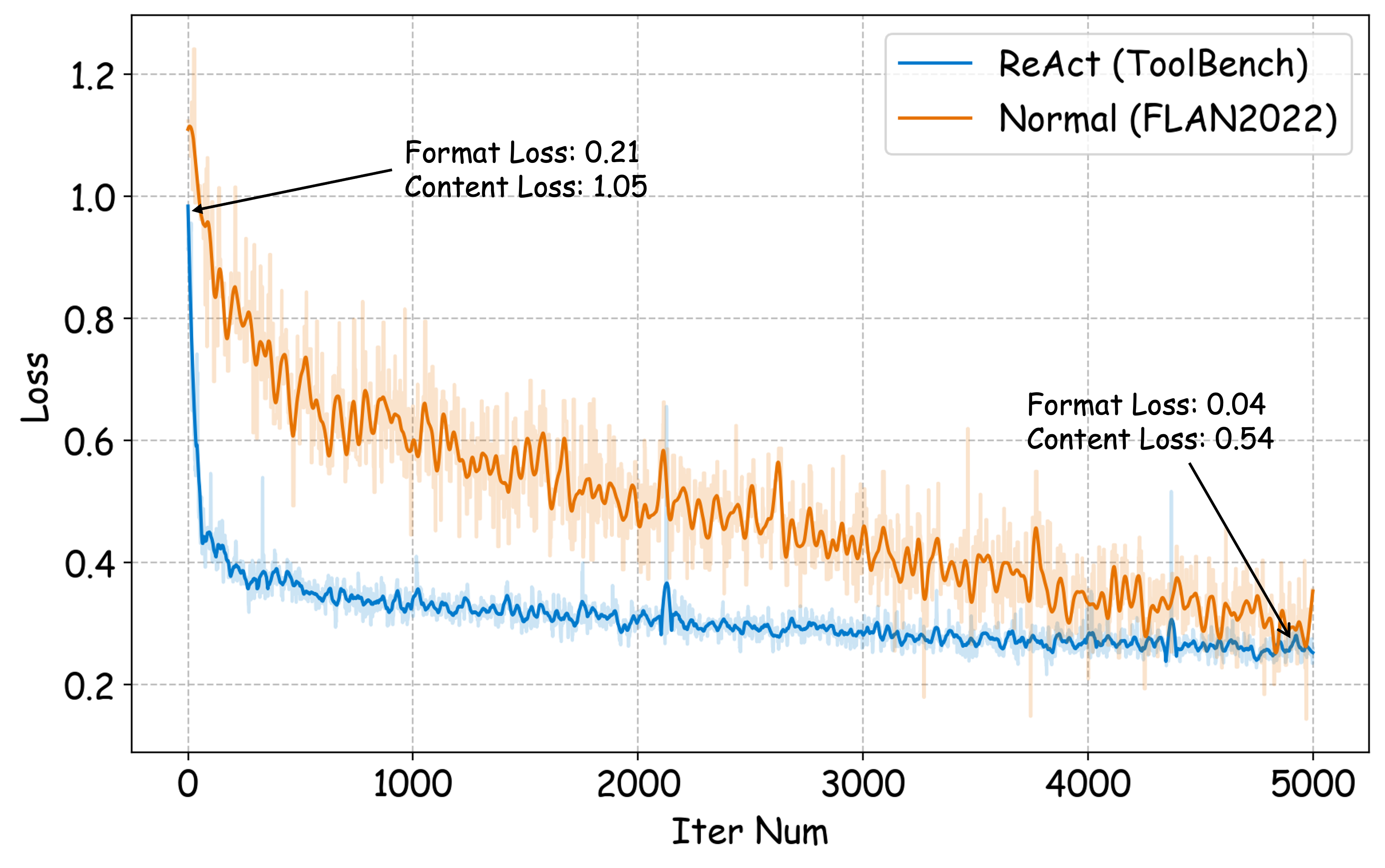}
    \vskip -0.5em
    \caption{Comparison of training loss on ReAct data (Toolbench \cite{qin2023toolllm}) and normal conversation (Flan2022 \cite{longpre2023flan}).}
    \label{fig:compare_loss}
\end{figure}
In this section, we delve into three pivotal observations on agent tuning that serve as the foundation of our subsequent investigation.\\
\textbf{Observation 1}. \textit{Most agent training data is entangled with both format following and general reasoning, causing a significant departure from the model's original pretraining language domain, namely, natural conversation.}\\
Recent agent tuning works \cite{zeng2023agenttuning,qin2023toolllm} endorse the adoption of specific formats, exemplified by ReAct (Thought-Action-Observation), for fine-tuning the language model. Moreover, it's noteworthy that action arguments are frequently presented in JSON format. Encoding both format and reasoning knowledge into the training corpus shifts the tuning process from the original chat domain, presenting it as an out-of-domain task for language models. As shown in Figure \ref{fig:compare_loss}, we compare the training curve of formatted data and normal data. It can be clearly seen that the loss associated with formatted data descends more rapidly to a low value, while keeping content loss still high (0.54 vs 0.04), indicating that the former leads to an inadequate learning process. This phenomenon can be probably attributed to the existence of a fixed structure (ReAct, JSON), where the model quickly gets overfitted to the format itself. Consequently, it fails to grasp the underlying reasoning abilities embedded within the training data, resulting in unsatisfied performance.

\begin{figure}[!h]
    \centering
    \includegraphics[width=.95\columnwidth]{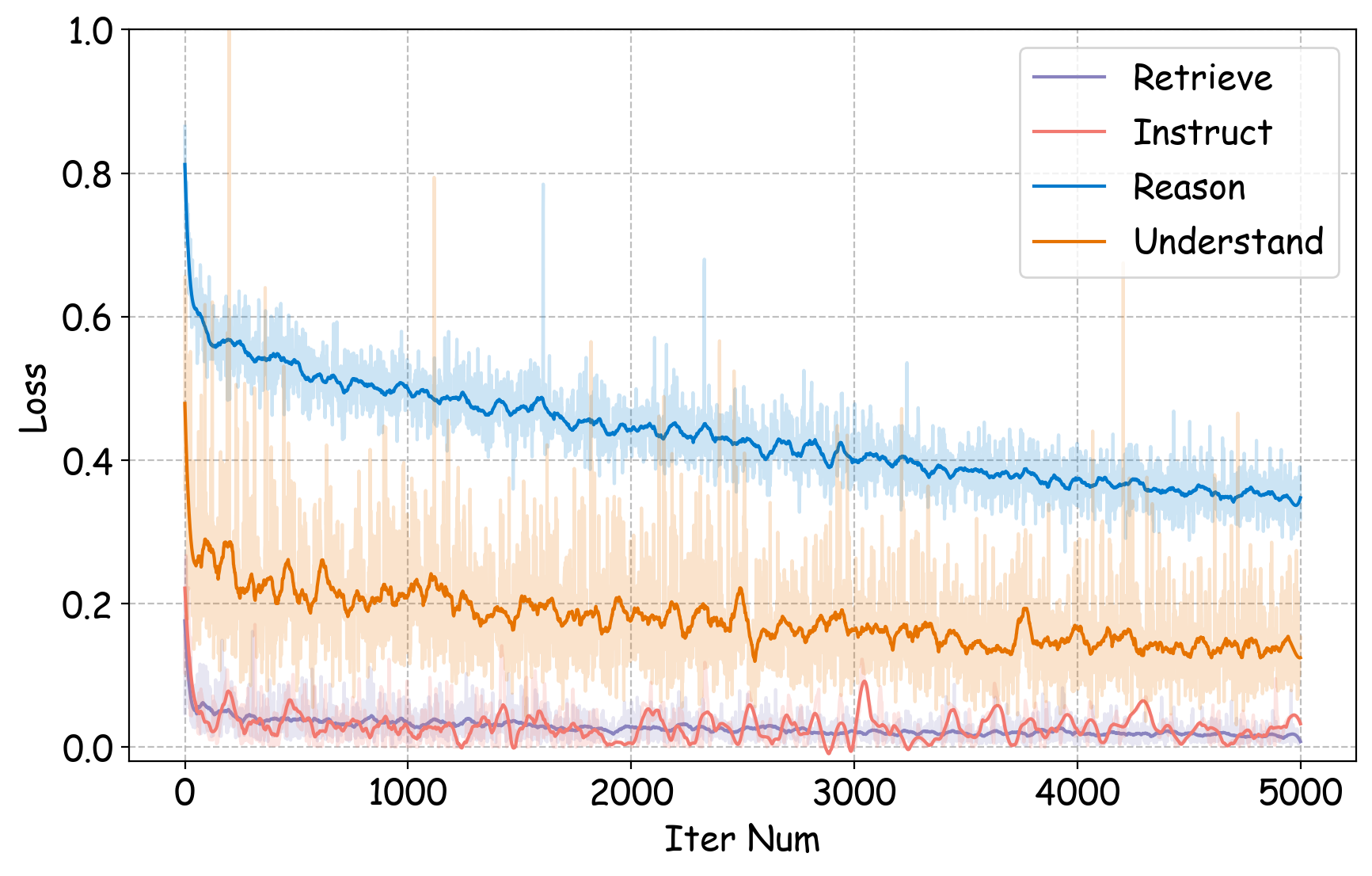}
    \vskip -0.5em
    \caption{Visualization of training loss by decomposing it into different capabilities of models: retrieval, instruct following, reasoning, and understanding.}
    \label{fig:multi_loss}
\end{figure}

\begin{figure*}[!t]
    \centering    
    \includegraphics[width=2.0\columnwidth]{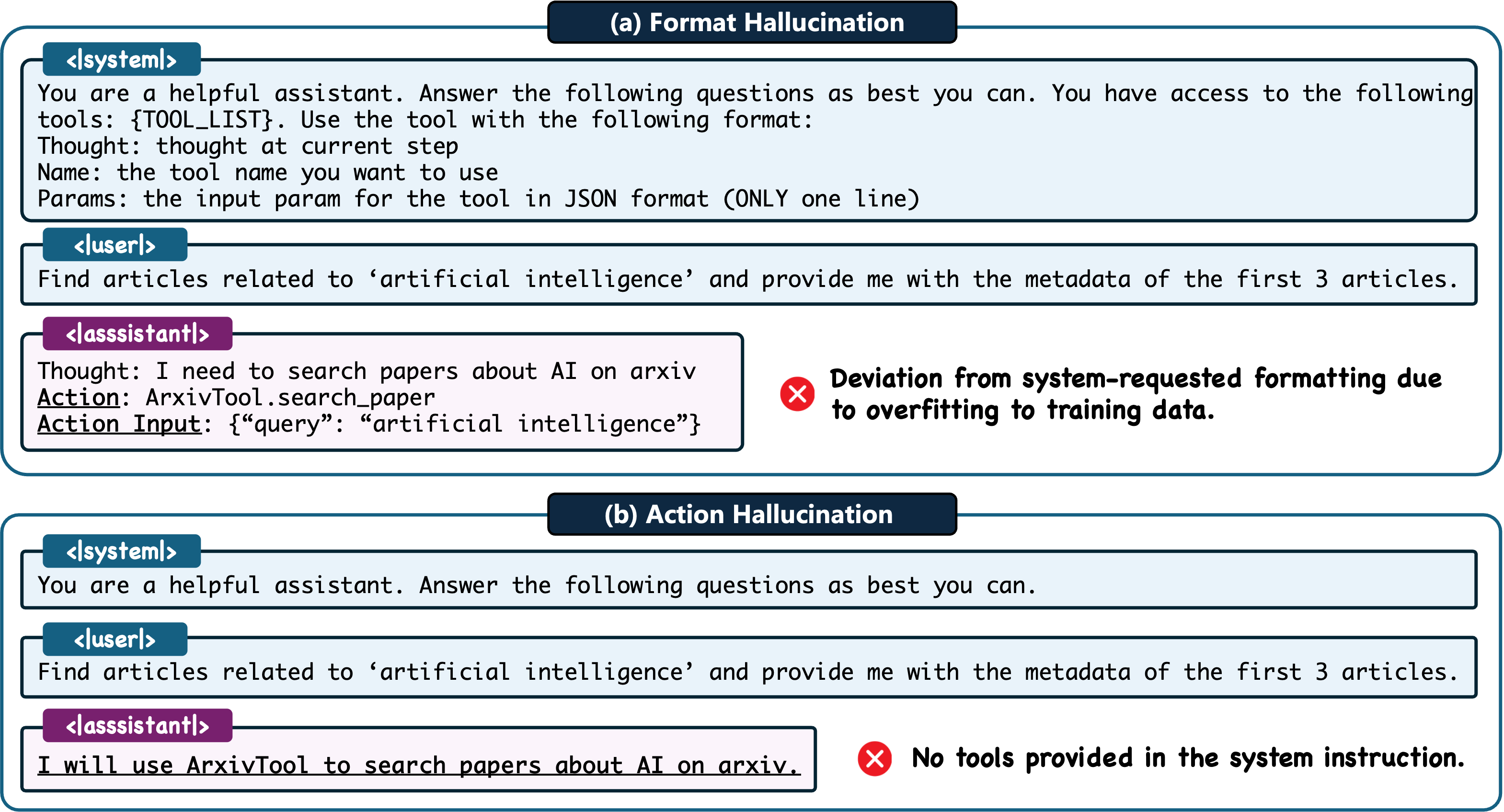}
    \vskip -0.5em
    \caption{Illustration of two typical hallucinations in general agent tasks for current open-sourced LLMs: (a) format hallucination and (b) action hallucination.}
    \label{fig:hallu}
\end{figure*}

\noindent\textbf{Observation 2}. \textit{By explicitly decomposing the training data along the basic capability aspects, each loss exhibits different convergence curves, indicating varied learning speeds on the capabilities requisite for agent tasks of LLMs}.\\
Inspired by \cite{chen2023t}, we explicitly disentangle the model's capabilities into distinct components: instruction following, reasoning, retrieval, and understanding. In this context, instruction following corresponds to format generation, reasoning corresponds to the thought quality at each step, retrieval involves selecting the appropriate function name to execute the task, and the understanding encompasses the parameter inputs for the selected functions. By visualizing the loss based on respective aspects in Figure \ref{fig:multi_loss}, we discern that LLM tends to exhibit varying learning speeds for the capabilities essential for proficient agents.
To elaborate, retrieval and understanding emerge as relatively more manageable tasks compared to reasoning, with instruction following being the simplest in the learning process. This observation serves as a compelling motivation to further disentangle the training data along these model capabilities and subsequently balance these data based on the model's varying learning rates.

\noindent\textbf{Observation 3}. \textit{Existing approaches predominantly concentrate on specialized agent abilities, overlooking the prevalence and significance of hallucination effects in the model's output.}\\
AgentTuning \cite{zeng2023agenttuning} introduces mixture training, where both a general dataset and an agent dataset are simultaneously provided to the model during the tuning process. While this strategy indeed leads to steady performance improvements, we observe that it has limited impact on addressing hallucination issues, \textit{i.e.,} a crucial concern often overlooked in recent agent studies. This concern becomes particularly significant when deploying language models in real-world applications, as illustrated in Figure \ref{fig:hallu}. 
Hallucination manifests in two primary aspects: (1) When the model is required to invoke a response, it strictly adheres to the training format, disregarding the user-generated query, and (2) The model is susceptible to being triggered with non-existent functions when presented with induced questions.
This underscores the necessity of directing more attention toward refining agent tuning mechanisms and the establishment of appropriate benchmarks to assess and mitigate agent hallucination effectively.

%
%

\section{Agent-FLAN}

Recent works start to explore the effectiveness of finetuning language models on various agent tasks, with varied data quality, model sizes, and tuning approaches. Building on the pilot observations, we dive deeper into the data and method designing choice of effective agent tuning and discuss three key improvements to the language agent models.

\begin{figure}[!t]
    \centering    
    \includegraphics[width=1.0\columnwidth]{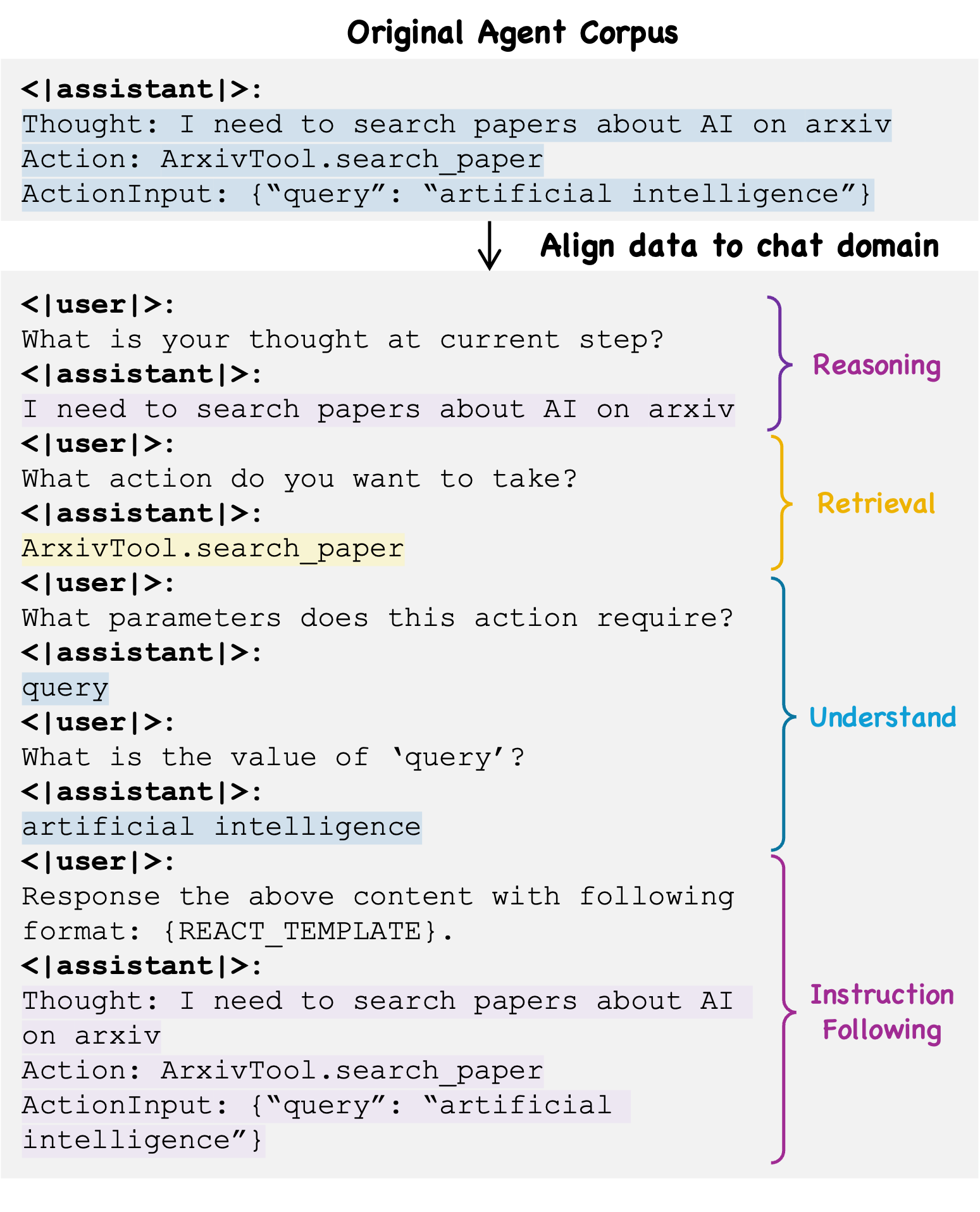}
    \vskip -0.5em
    \caption{By aligning the original agent corpus to the natural conversation, we are able to explicitly decompose agent tasks into different capabilities, leading to more fine-grained data balancing.}
    \label{fig:align_agent}
\end{figure}

\begin{table*}[t]
    \small
    \centering
     \caption{\textbf{Main results of Agent-FLAN}. Agent-FLAN significantly outperforms previous agent-tuning approaches by a large margin on both held-in and held-out tasks. * denotes our re-implementation with the same amount of training data for a fair comparison. Since FireAct does not train on AgentInstruct dataset, we omit its performance on the \textsc{Held-In} set. \textbf{Bold}: the best in API-based and open-sourced models.}
    \label{tab:ablations}
    \vskip 0.75em
    \begin{tabular}{l|c|ccccc|c}
    \toprule
    \multirow{2}{*}{\textsc{\textbf{Model}}} & \multirow{2}{*}{\textsc{\textbf{Held-In}}} & \multicolumn{6}{c}{\textsc{\textbf{Held-Out}}}\\
     & & HotpotQA  & SciWorld & WebArena & T-Eval & Agent-H & Overall \\
    \midrule
    GPT-3.5 \cite{opeiai2022gpt} & 1.59 & 37.4 & 21.2 & 4.56 & 84.0 & 92.1 & 47.8  \\
    GPT-4 \cite{openai2023gpt4} & \textbf{2.75} & \textbf{52.1} & \textbf{36.4} & \textbf{6.28} & \textbf{86.4} & \textbf{94.2} & \textbf{55.1}\\
    \midrule
    Llama2-7B \cite{touvron2023llama} & 0.19 & 22.6 & 5.9 & 1.2 &  27.4 & 78.7 & 27.2\\
    FireAct-7B \cite{chen2023fireact} & - & 26.2 & 6.8 & 0.25 & 9.3 & 40.4 & 16.6\\
    AgentLM-7B \cite{zeng2023agenttuning} & 1.96 & 22.3 & 13.7 & 0.74 & 41.4 & 80.6 & 31.7\\
    \midrule
    AgentTuning* \cite{zeng2023agenttuning} & 1.89 & 25.4 & 16.8 & 2.71 & 61.8 & 84.5 & 38.2\\
    Agent-FLAN (Ours) & \textbf{2.01} & \textbf{28.5} & \textbf{20.0} & \textbf{4.68} & \textbf{66.0} & \textbf{89.1} & \textbf{41.7}\\
    \bottomrule
    \end{tabular}
\end{table*}

\noindent\textbf{Experimental Setup} We finetune the language model Llama2-series \cite{touvron2023llama}, and use 7B size for ablations for efficiency unless otherwise stated. We construct the dataset in accordance with the data and settings established by AgentTuning \cite{zeng2023agenttuning}. Specifically, we pick a suite of training data sources as held-in tasks: ALFWorld \cite{shridhar2020alfworld}, WebShop \cite{yao2022webshop}, Mind2Web \cite{deng2023mind2web}, Knowledge Graph \cite{liu2023agentbench}, Operating System \cite{liu2023agentbench}, Database \cite{liu2023agentbench}, and ToolBench \cite{qin2023toolllm}, covering both general agents and tool utilization domains. Our held-out evaluation encompasses a list of complex interactive tasks, including complex QA (HotpotQA \cite{yang2018hotpotqa}), web browsing (WebArena \cite{zhou2023webarena}), science experiments (SciWorld \cite{wang2022scienceworld}), and tool utilization (T-Eval \cite{chen2023t}). Details adopted in Agent-FLAN and hyper-parameters during training can be found in Appendix \ref{app:exp_setup}.

\subsection{Aligning Agent Tuning to Pretrain Domain}
\label{sec:method_align}

LLMs are first pretrained on natural language corpus, which comprises a massive number of human conversations. However, agent data are often presented in specific formats (ReAct, JSON), leading to out-of-distribution learning during the tuning process. Such a misalignment further results in an inadequate learning process. Besides, LLMs are more likely to get overfitted to these specific formats after fine-tuning, deteriorating their instruction-following abilities. To mitigate this problem, we propose to transform the formatted data into natural conversations. Concretely, we first replace the classical `\textsc{Thought-Action-ActionInput}' templates with multi-turn dialogues. After that, we further decompose JSON arguments by inserting several elicit statements. An example has been shown in Figure \ref{fig:align_agent}. Since loss will be only applied to the `assistant' clause, the introduction of formatted elicit statements has little overfitting issue on models. By explicitly aligning the agent corpus into chat domain, we are able to fully boost the learning on pure agent ability, without focusing on strict format protocols. In order to keep the ability to output various requested formats, we additionally construct instruction following pairs which request the model to respond with ReAct and JSON formats. Our latter experiments also show that only a small portion of instruct following data is enough to achieve satisfying results. From Table \ref{tab:mixture-ranking}, we can observe steady improvements by aligning training corpus into chat domain, \textit{i.e.,} 3.1\% improvements on T-Eval, and 2.5\% on HotpotQA. This further validates the correctness and effectiveness of the alignment of training corpus to the chat format.

\begin{table}[!t]
    \small
    \centering
    \caption{Subsets of decomposed capabilities are left out with half from an equally weighted mixture to measure their importance.}
    \vskip 0.75em
    \label{tab:mixture-ranking}
    \begin{tabular}{l | c| ccc}
    \toprule
    Methods & Tokens (M) & T-Eval & HotpotQA\\
    \midrule
    ReAct (Baseline) & 19.2 & 61.8 & 25.4\\
    Align-Chat (All) & 37.3 & 64.9 & 27.9\\
    \midrule
    All - Reasoning & 32.4 & 63.8 & 27.4\\
    All - Retrieval & 36.2 & 65.3 & \textbf{29.0}\\
    All - Understand & 35.4 & 64.6 & 28.1\\
    All - Inst. & 28.4 & 65.9 & 27.5\\
    \midrule
    All (Weighted) & \textbf{18.1} & \textbf{66.3} & 28.5\\
    \bottomrule
    \end{tabular}
    
\end{table}


\subsection{Capabilities Decomposition and Data Balancing}
\label{sec:method_decompose}

Prior works have shown that a proper mixture of training data sources leads to better performance \cite{longpre2023flan}. In this work, instead of simply exploring the balanced composition of each dataset, we investigate the mixture of training corpus from the perspective of capabilities. Inspired by \cite{chen2023t}, we explicitly decompose the agent data along the capabilities required by each task, including reasoning, retrieval, understanding, and instruction following. As demonstrated in Section \ref{sec:pilot}, LLM exhibits varied learning speeds on each capability, indicating that properly composing these data sources also matters in optimizing final results. 
To validate this assumption, we conduct experiments with vanilla decomposed aligned agent corpus, where each ability data is identical to the original data, and then ablates each sub-mixture with half of the data on T-Eval and HotpotQA. 

As shown in Table \ref{tab:mixture-ranking}, reasoning and understanding are among the most beneficial mixtures, then comes retrieval and instruction following: when reducing the portion of reasoning and understanding data to 50\%, the final performance declines 1.1/0.3 points, respectively. By reducing the amount of retrieval and instruction following, the performance incurs little influence, and even improves. These findings are also consistent with the observations in Figure \ref{fig:multi_loss}, where the loss of retrieval and instruction following drops much quicker than reasoning and understanding. Such a phenomenon also inspires us to narrow the mixture weights search space and greatly reduce the training tokens based on the loss curves on each capability. 

\begin{figure}[!t]
    \centering    
    \includegraphics[width=1.0\columnwidth]{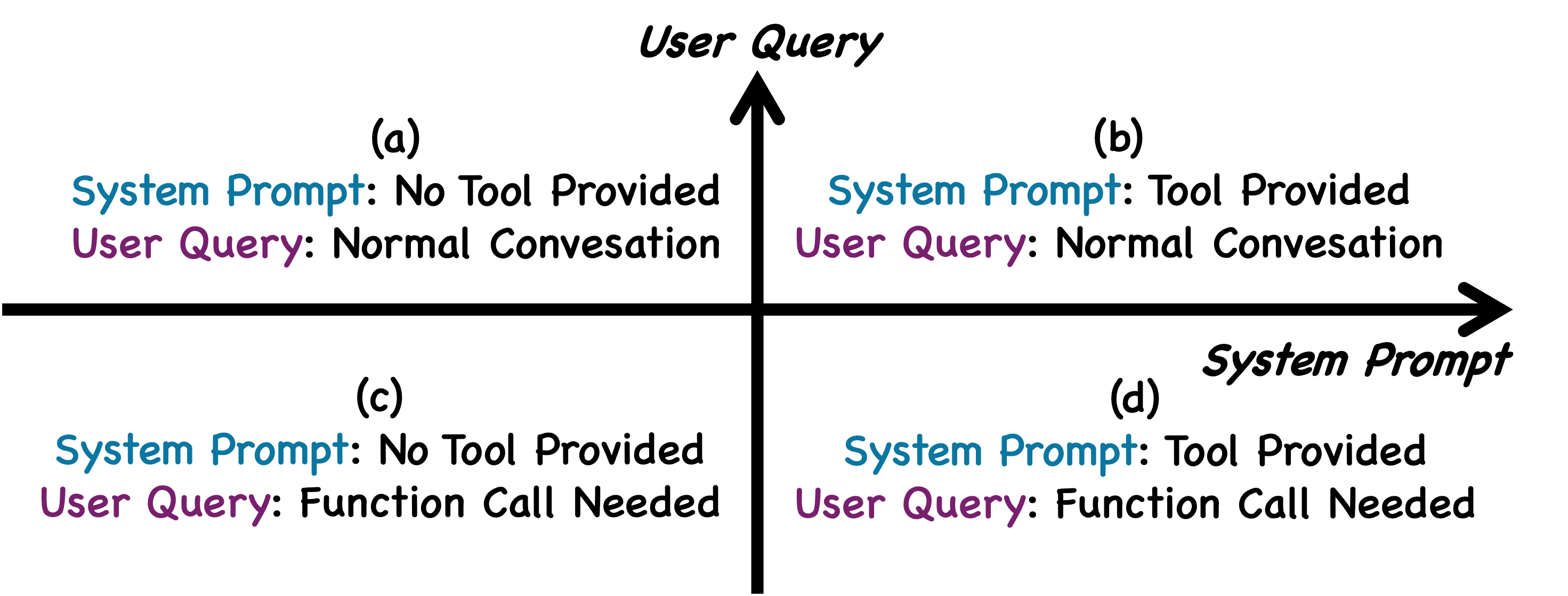}
    \vskip -0.5em
    \caption{Demonstration of 4 different circumstances commonly existed in agent tasks, by splitting them through user query and system prompt.}
    \label{fig:hallu_cls}
    \vskip -1em
\end{figure}

\subsection{Negative Sample Learning for Hallucination Elimination}
\label{sec:method_negative}

Hallucination is a crucial issue of current LLMs, which refers to ``generating unfaithful or nonsensical text'' \cite{ji2023survey}. In agent tasks, we also observe such phenomenons as illustrated in Figure \ref{fig:hallu}. We summarize agent hallucination into two main categories: format hallucination and action hallucination. The former plays an important role when deploying LLMs into specific agent systems, and the latter matters when acting as a general chat assistant. Therefore, how to effectively eliminate the agent hallucination issues is an essential path in developing agent LLMs. 

However, most prior work mainly focuses on the general agent's abilities, while omitting the hallucination issues. To comprehensively gauge the LLMs on agent hallucination, we first establish \textit{Agent-H} benchmark, which accesses such issues from two aspects: (1) format-level: requests with various response formats, and check if the model follows the instructions, and (2) action-level: we curate the questions from 4 different perspectives as illustrated in Figure \ref{fig:hallu_cls}, covering most agent circumstances for LLMs. 

Specifically, we select glaive-function-calling-v2 \cite{glaiveai2023glaive} as our base dataset. By explicitly checking if the response contains tool invocation, we curate 1845 samples for out-of-domain validation. Since we focus on the hallucination issues when acting as an agent, the evaluation protocol only gauges if the output of the model is a raw response or specified function calling. Specifically, we define two specific format checkings: (1) ReAct-format hallucination (\textit{e.g.,} `Thought:', `Action:'), and (2) general-format hallucination (\textit{e.g.,} `I will use', `I need to call'). If the response contains the above keywords, it will be viewed as one hallucination failure when the ground truth is a raw response type. Based on these, we further define two numerical metrics: H$_\text{ReAct}$ and H$_\text{General}$ as the number of respective format hallucinations / number of raw responses in the ground truth. The final overall score H$_\text{Score}$ is a reverse average of the above two metrics:
\begin{equation}
    \text{H}_\text{Score} = 0.5 * ((1 - \text{H}_\text{ReAct}) + (1 - \text{H}_\text{ReAct}))
\end{equation}
Table \ref{tab:agent_h} reports the experimental results on \textit{Agent-H}, as well as the scores on T-Eval, which provides a comprehensive demonstration of both agent abilities and hallucination issues. Ideally, a general language model should obtain high scores on both benchmarks. From the table, we can see that Llama2-7B obtains low scores on both \textit{Agent-H} and T-Eval.
This is possibly due to the lack of agent data in its pretraining corpus, which further proves the necessity of agent tuning. We also follow the implementation of AgentTuning \cite{zeng2023agenttuning} to finetune the model from Llama2-7B. Despite the huge improvements in T-Eval score, the hallucination issue is quite severe gauged by \textit{Agent-H}, pinpointing the inner defects of current agent tuning approaches. 

To address this problem, we first examine the current agent corpus according to Figure \ref{fig:hallu_cls}. It is easy to find that in most cases, the training data only covers normal conversation without tools (a) and agent tasks with provided tools (d) while omitting the rest (b,c). Since the model has never seen these negative samples during training, it can hardly generalize to these requests, leading to unexpected responses. To this end, we introduce negative sample learning, by meticulously curating diverse negative training samples covering various conditions mentioned above. Specifically, we insert two different types of negative samples: (1) no tools provided, user query requests for tools (2) tool provided, user query requests for normal conversations. Through explicit supervision, we teach the model not only \textit{how} but \textit{when} to act as an agent. In Table \ref{tab:agent_h}, the negative sampling learning strategy significantly mitigates the hallucination issues, meanwhile maintaining the high performance on T-Eval.

\begin{table}[!t]
    \small
    \centering
    \caption{Experimental results on Agent-H with Llama2-7B. H$_\text{score}$ is the overall score of Agent-H benchmark. ``NS'' denotes negative training samples.}
    \vskip 0.75em
    \label{tab:agent_h}
    \begin{tabular}{c | c | cc|c}
    \toprule
    \multirow{2}{*}{Method} & \multirow{2}{*}{T-Eval} & \multicolumn{3}{c}{Agent-H}\\
    &  & H$_\text{ReAct}\downarrow$ & H$_\text{General}\downarrow$ & H$_\text{Score}\uparrow$ \\
    \midrule
    Llama2-7B & 27.4 & 21.7 & 21.0 & 78.7 \\
    AgentTuning & 61.8 & 18.1 & 14.0 & 83.9\\
    \midrule
    Agent-FLAN & 66.0 & \textbf{9.9} & \textbf{11.9} & \textbf{89.1}\\
    w/o NS & \textbf{66.3} & 15.6 & 13.5 & 84.5\\
    \bottomrule
    \end{tabular}
    
\end{table}

\section{Analysis}

In this section, we conduct in-depth experiments to analyze  
\subsection{Scaling Law for Agent Tuning}

Scaling law for language models is quite important and provides valuable insights in further development \cite{longpre2023flan,chung2022scaling}. In this section, we explore such phenomenons on agent tuning from both the data and model scales on the general HotpotQA task.

\subsubsection{Data Scaling Law for Agent Tuning} Recent work \cite{chung2022scaling} has demonstrated language models can benefit from vast and diverse training corpus in general abilities. We investigate how the amount of training data influences the agent's abilities. By evenly dividing Agent-FLAN data into 25\%, 50\%, 75\%, and 100\%, we report results in Figure \ref{fig:data_scale}. It can be seen that with only 25\% training samples, the agent ability gains the most. This further verifies that the vanilla Llama-2 model exhibits weak agent abilities and requires specific training, and only a small portion of agent corpus can elicit most abilities required by agent tasks. When further increasing the data amount (50\%, 75\%), the improvements persist, however with slower speeds, which indicates that simply enlarging the scale of the training corpus of agents does not contribute much to the model capabilities. Therefore, enriching the diversity or improving the quality of the training corpus may be the necessary path to better language agents.

\begin{figure}[!h]
    \centering    
    \includegraphics[width=0.95\columnwidth]{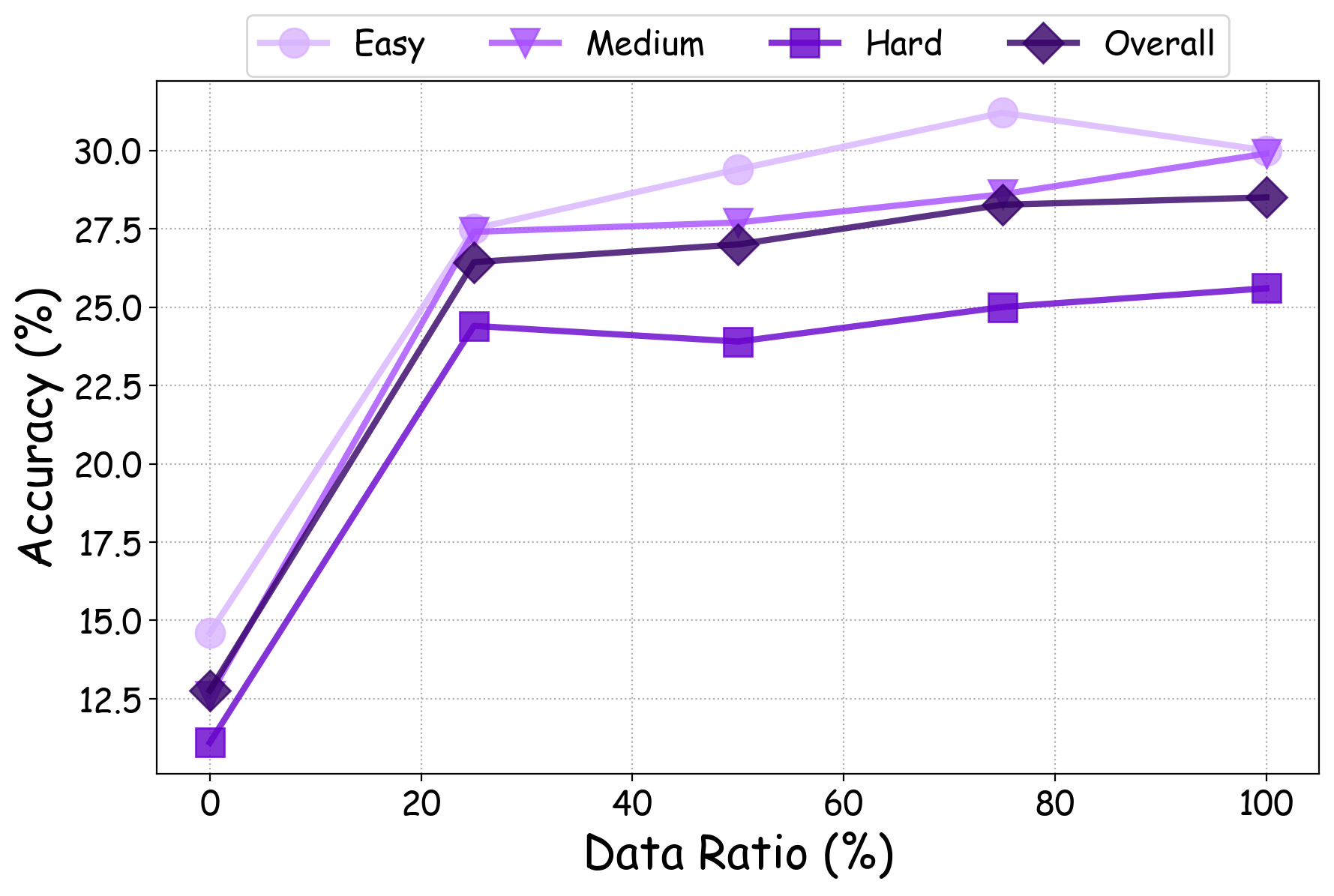}
    \vskip -0.5em
    \caption{Performance scaling laws for the amount of training data, ranging from 0\% to 100\%.}
    \label{fig:data_scale}
\end{figure}

\subsubsection{Model Scaling Law for Agent Tuning} As previous work \cite{longpre2023flan} proves that zero/few-shot ability of language models is substantially improved with a larger model scale, we next explore if this rule is also applied in the agent domain. By evaluating our approach on models of size 7B, 13B, and 70B on Llama2, the final results are shown in Figure \ref{fig:model_scale}. We can observe that as the model scale increases, the performance continuously improves without any saturation, proving that larger parameters do guarantee better performances. 
When taking a close look at enhancement across different model scales, it can be found that specific agent tuning brings steady improvements compared to vanilla ReAct tuning as the model scales up. We infer the reason that larger models have already possessed basic abilities required by agent tasks, such as reasoning and retrieval. Therefore, it is more important to elicit agent capability with a certain amount of agent tuning corpus in an appropriate way. 

\begin{figure}[!h]
    \centering    
    \includegraphics[width=0.95\columnwidth]{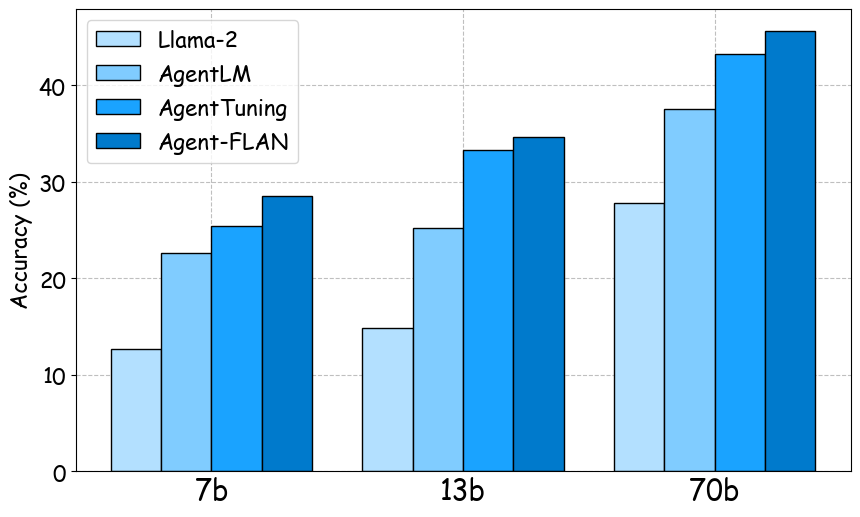}
    \vskip -0.5em
    \caption{Performance scaling laws for the parameters of training models, from Llama2-7B to Llama2-70B. 
    }
    \label{fig:model_scale}
\end{figure}

\begin{figure*}[!t]
    \centering    
    \includegraphics[width=2.0\columnwidth]{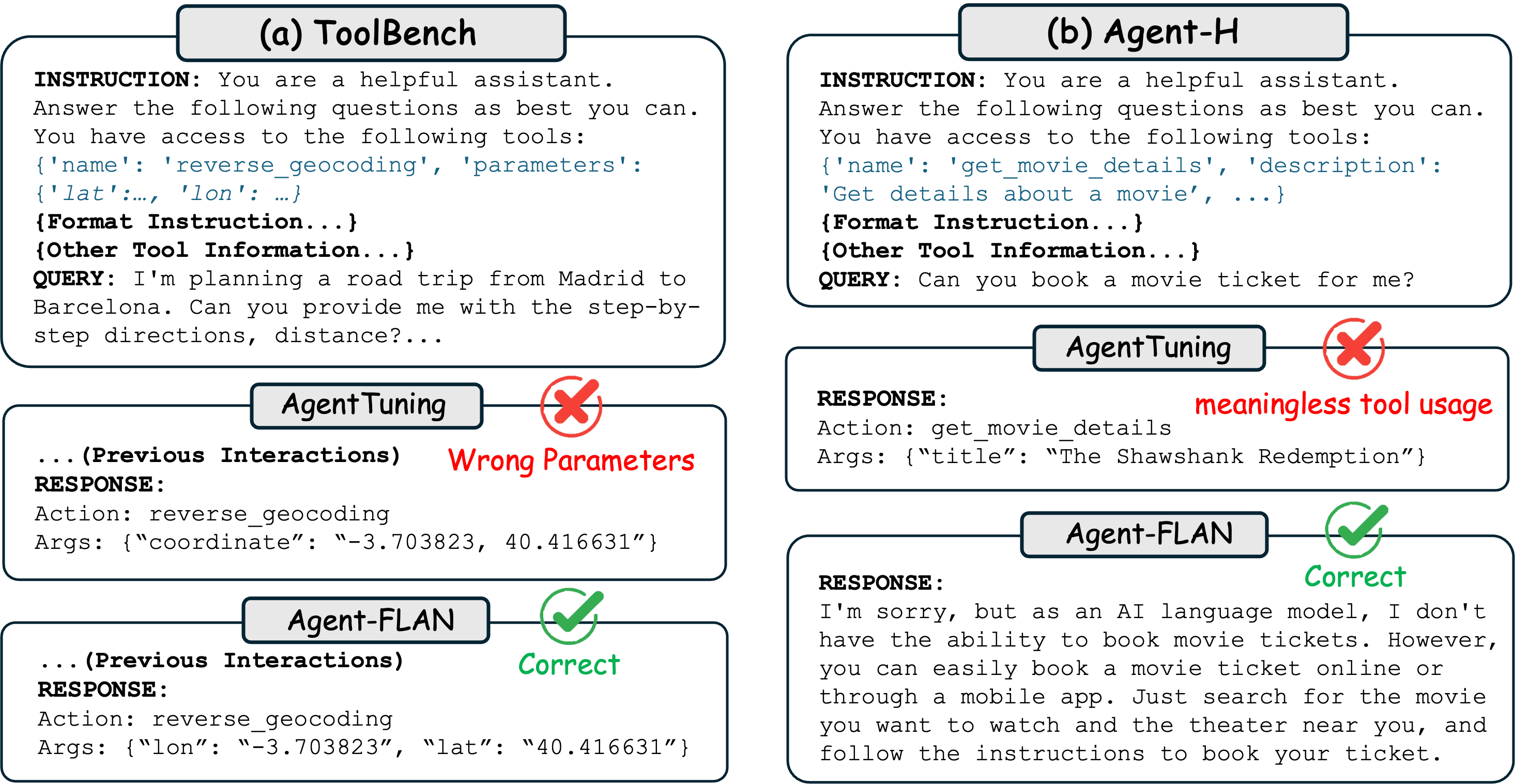}
    \caption{\textbf{Comparison studies on Toolbench and Agent-H datasets between AgentTuning and Agent-FLAN with Llama2-7B}. (a) ToolBench: Thanks to the capability decomposition and more focus tuning on `understand', Agent-FLAN is able to catch up with the specific API information given long tool information content, whereas AgentTuning failed with hallucination. (b) Agent-H: the AgentTuning model presents a meaningless tool usage while Agent-FLAN directly gives the preferred response.}
    \label{fig:case_study}
\end{figure*}

\subsection{General Ability v.s Agent Ability}

As we have seen that specific tuning improves the ability of a model to act as agent, it is still unclear what is the relationship between general ability and agent ability. AgentTuning has verified the necessity of general data on agent training, but does agent tuning further promote the general capability of the model? We evaluate our model on three general capabilities that are widely adopted in the field: MMLU (linguistic knowledge), GSM8K (mathematical ability), and HumanEval (code capability). The results are shown in Table \ref{tab:general}. From the table, we can see that introducing agent training corpus not only enhances the ability on agent tasks but also brings extra benefits to general capabilities. We speculate the reason that agent corpus contains basic capabilities such as reasoning, and instruction following, which are also amenable to other domains. This further indicates that properly integrating agent data into the current LLM training corpus can lead to better performance.

\definecolor{newg}{RGB}{46, 184, 46}
\definecolor{newp}{RGB}{76, 0, 153}
\begin{table}[!h]
    \small
    \centering
    \caption{Effectiveness of AgentFLAN corpus on the general capabilities with Llama2-series.}
    \vskip 0.75em
    \label{tab:general}
    \begin{tabular}{c | lll}
    \toprule
    \multirow{2}{*}{\textsc{Data}} & \multicolumn{3}{c}{\textit{Llama2-7B}} \\
     & \textsc{MMLU} & \textsc{GSM8K} & \textsc{HumanEval} \\
    \midrule
    general data & \textbf{50.0} & 21.9 & 15.1\\
    Agent-FLAN & 49.7 \small{\textcolor{newp}{-0.3}} & \textbf{22.1} \small{\textcolor{newg}{+0.2}}  & \textbf{15.5} \small{\textcolor{newg}{+0.4}}\\
    \midrule
    \multirow{2}{*}{\textsc{Data}} & \multicolumn{3}{c}{\textit{Llama2-13B}} \\
     & \textsc{MMLU} & \textsc{GSM8K} & \textsc{HumanEval} \\
    \midrule
    general data & 54.7 & 34.8 & 15.2\\
    Agent-FLAN & \textbf{55.8} \small{\textcolor{newg}{+1.1}} & \textbf{35.2} \small{\textcolor{newg}{+0.4}} & \textbf{15.8} \small{\textcolor{newg}{+0.6}}\\
    \midrule
    \multirow{2}{*}{\textsc{Data}} & \multicolumn{3}{c}{\textit{Llama2-70B}} \\
     & \textsc{MMLU} & \textsc{GSM8K} & \textsc{HumanEval} \\
    \midrule
    general data & 68.0 & 64.5 & 32.1\\
    Agent-FLAN & \textbf{68.5} \small{\textcolor{newg}{+0.5}} & \textbf{64.6} \small{\textcolor{newg}{+0.1}}& \textbf{32.9} \small{\textcolor{newg}{+0.8}}\\
    \bottomrule
    \end{tabular}
    
\end{table}
\section{Conclusion}

In this work, we investigate the design choice of data and methods in agent tuning for LLMs. By first carrying out three critical observations, we pinpoint the current bottleneck in bridging the gap between the open-sourced LLMs and API-based models. Based on this, we present Agent-FLAN to effectively fine-tune the language models for agent tasks. Through careful decomposition and redesign of the existing training corpus, Agent-FLAN empowers Llama2-7B to outperform previous works significantly on a wide spectrum of agent tasks. 

\section{Limitations}

In this paper, we focus on constructing the agent training corpus. Despite our best efforts, this paper may still have some remaining limitations. 1) The training and validation dataset only encompass part of agent tasks. There are many other interactive scenarios beyond this. We will conduct further research on applying Agent-FLAN to a wider range of benchmarks in the future. 2) In order to keep the quality of training data, we only choose roughly 20,000 valid samples from ToolBench, which is 10\% of the whole dataset. It would be better to fully utilize them to further enhance the performance of the model. We leave it for future work.

\section{Ethical Considerations}
We used publicly available reference documents/APIs for our experiments, effectively circumventing any possible harm toward individuals or groups.
The generated data by LLMs were carefully selected and processed by humans to secure privacy and confidentiality. No personal identification information was involved, and all data were made anonymous before any analysis was conducted. 
Besides, We use ChatGPT and Grammarly to polish the writing.

\nocite{langley00}

\newpage

\bibliography{main.bib}
\bibliographystyle{icml2021}

\clearpage
\definecolor{lightgray}{gray}{0.95}
\lstdefinestyle{prompt}{
    basicstyle=\ttfamily\fontsize{7pt}{8pt}\selectfont,
    frame=none,
    breaklines=true,
    backgroundcolor=\color{lightgray},
    breakatwhitespace=true,
    breakindent=0pt,
    escapeinside={(*@}{@*)},
    numbers=none,
    numbersep=5pt,
    xleftmargin=5pt,
}
\tcbset{
  aibox/.style={
    top=10pt,
    colback=white,
    colframe=black,
    colbacktitle=black,
    enhanced,
    center,
    attach boxed title to top left={yshift=-0.1in,xshift=0.15in},
    boxed title style={boxrule=0pt,colframe=white,},
  }
}

\appendix
\section{Experimental Details}
\label{app:exp_setup}
For Agent-FLAN experiments, we follow the practice in AgentTuning, where ShareGPT and agent corpus are mixed for training with a mix ratio of 1:1. Besides, we empirically find that a small amount of ReAct format data improves the performance. Therefore, we incorporate 10\% data with ReAct format and the rest follows the format in Agent-FLAN. The models are fine-tuned on the training corpus with 1 epoch. We list the hyper-parameters of models with different scales in Table \ref{tab:hyper}. The balancing weight for reasoning, retrieval, and understanding is set to 1:0.25:0.75. We additionally curate 2000 samples of instruction following data for training.

\begin{table}[!h]
    \small
    \centering
    \caption{Experimental settings and hyper-parameters for Llama2-series.}
    \vskip 0.75em
    \label{tab:hyper}
    \begin{tabular}{c | c c c }
    \toprule
    Hyper-Params & Llama2-7B & Llama2-13B & Llama2-70B\\
    \midrule
    GPUs & 16 & 32 & 128\\
    Learning Rate & 2e-5 & 2e-5 & 2e-5\\
    WarmUp LR & 4e-6 & 4e-6 & 4e-6\\
    Batch Size & 32 & 32 & 128\\
    LR Schedule & Cosine & Cosine & Cosine \\
    \bottomrule
    \end{tabular}
    
\end{table}

\section{Dataset Details}

In this section, we provide details on the detailed construction of our constructed Agent-FLAN dataset. In order to keep the generality of our approach, we adopt open-sourced datasets AgentInstruct \cite{zeng2023agenttuning} (which comprises 6 different agent tasks), ToolBench \cite{qin2023toolllm} as our basic datasets. We filter the data that does not follow the format of `Thought-Action-ActionInput', and remove the few-shot examples in the AgentInstruct. As for ToolBench dataset, we only keep samples that have \textsc{FinalAnswer} and remove those that directly begin with \textit{`This is not the first time you try this task'} (due to the DFSDT introduced in \cite{qin2023toolllm}). By default, Agent-FLAN remixes 10\% ReAct format and 90\% conversation format data for better performance. The overview of our dataset is shown in Table \ref{tab:aflan_details}. 

\begin{table}[!h]
    \small
    \centering
    \caption{Dataset statistics used in Agent-FLAN.}
    \vskip 0.75em
    \label{tab:aflan_details}
    \begin{tabular}{c | c }
    \toprule
    Datasets &  Sample Num \\
    \midrule
    OS & 195 \\
    Database & 538 \\
    Knowledge Graph & 300 \\
    Webshop & 351 \\
    AlfWorld & 336 \\
    Mind2Web & 116 \\
    ToolBench & 22,867 \\
    \midrule
    Total & 24,703\\
    \bottomrule
    \end{tabular}
    
\end{table}

%


\section{Negative Samples Construction Details}
In order to construct negative samples for hallucination elimination, we first extract a small portion (761 samples) of user queries from ToolBench dataset. Then we prompt the \texttt{gpt-3.5-turbo} to directly reply to these queries without providing any tool information. To construct normal conversation with the tool information provided, we randomly append half of the queries with irrelevant tools, which generate the case (b) in Figure \ref{fig:hallu_cls}. Illustrations of our constructed negative samples are shown in Figure \ref{fig:ns1}, \ref{fig:ns2}.

\newtcolorbox{AIbox}[2][]{aibox, title=#2,#1}
\begin{figure*}[!ht] 
\vspace{-5mm}
\begin{AIbox}{Negative Sample I}
{\color{blue}\bf System:} \\
{
You are a helpful assistant.
}\\
{\color{orange}\bf User Query:} \\
{
I'm a crypto blogger and I'm writing an article about The Sandbox - SAND. Can you provide me with the price data for the past 1 week, 1 year, and 3 months? It would be great to include the market caps and total volumes as well.
}\\
{\color{teal}\bf Assistant:} \\
{
As an AI language model, I don't have real-time data. However, you can easily find the price data for The Sandbox (SAND) on various cryptocurrency tracking websites like CoinMarketCap or CoinGecko. These platforms provide historical price data, market caps, and total volumes for different time periods.
}
\end{AIbox} 
\caption{An example of constructed negative samples (I).}
\label{fig:ns1}
\end{figure*}

\begin{figure*}[!ht] 
\vspace{-5mm}
\begin{AIbox}{Negative Sample II}
{\color{blue}\bf System:} \\
{
You are a helpful assistant. You can use the following tools: 
\begin{lstlisting}[style=prompt]
[
    {
        'name': 'get_recipe', 
        'description': 'Get a recipe based on given ingredients', 
        'parameters': {'type': 'object', 'properties': {'ingredients': {'type': 'array', 'items': {'type': 'string'}, 'description': 'The list of ingredients available'}}, 
        'required': ['ingredients']}
    }, 
    {
        'name': 'convert_weight', 
        'description': 'Convert weight from one unit to another', 
        'parameters': {'type': 'object', 'properties': {'weight': {'type': 'number', 'description': 'The weight value'}, 'from_unit': {'type': 'string', 'description': 'The unit to convert from'}, 'to_unit': {'type': 'string', 'description': 'The unit to convert to'}}, 
        'required': ['weight', 'from_unit', 'to_unit']}
    }
]
\end{lstlisting}
}
{\color{orange}\bf User Query:} \\
{
I am planning a big event for my company and I need to minimize costs. Can you assist me in finding the latest coupons, trending coupons, and popular coupons for event venues, catering services, and transportation? Please provide the coupon codes, discount percentages, and expiration dates for each coupon.
}\\
{\color{teal}\bf Assistant:} \\
{
I'm sorry, but as an AI language model, I don't have real-time access to current coupons or their details. I recommend checking popular coupon websites or contacting specific event venues, catering services, and transportation providers directly for any available discounts or promotions.
}
\end{AIbox} 
\caption{An example of constructed negative samples (II).}
\label{fig:ns2}
\end{figure*}

%
%
%

\end{document}